%% file: neurips_2025.tex
\documentclass{article}

\PassOptionsToPackage{numbers, compress}{natbib}

\usepackage[preprint]{neurips_2025}

\usepackage[utf8]{inputenc} 
\usepackage[T1]{fontenc}    
\usepackage{hyperref}       
\usepackage{url}            
\usepackage{booktabs}       
\usepackage{amsfonts}       
\usepackage{nicefrac}       
\usepackage{microtype}      
\usepackage{xcolor}         
\usepackage{amsmath}
\usepackage{graphicx}
\usepackage{wrapfig}
\usepackage{multirow}
\usepackage{threeparttable}
\usepackage{float}
\usepackage{amssymb}
\usepackage[ruled,vlined]{algorithm2e}
\usepackage{stmaryrd}
\usepackage{booktabs}
\usepackage{subcaption}
\usepackage{pifont}
\usepackage{wrapfig}

\usepackage[nameinlink, capitalise]{cleveref}

\title{Cognitive Structure Generation: From Educational Priors to Policy Optimization}

\author{%
  Hengnian Gu$^{\dag}$ \quad Zhifu Chen$^{\dag}$ \quad Yuxin Chen$^{\dag}$ \quad Jin Peng Zhou$^{\ddag}$ \quad Dongdai Zhou$^{\dag}$\thanks{Corresponding author.} \\
  $^{\dag}$Northeast Normal University, Changchun, Jilin, China\\
  $^{\ddag}$Cornell University, Ithaca, New York, United States\\
  \texttt{\{guhn546,zhifuchen,chenyx395,ddzhou\}@nenu.edu.cn, jpzhou01@gmail.com}
}

\begin{document}

\maketitle

\begin{abstract}
  Cognitive structure is a student's subjective organization of an objective knowledge system, reflected in the psychological construction of concepts and their relations. However, cognitive structure assessment remains a long-standing challenge in student modeling and psychometrics, persisting as a foundational yet largely unassessable concept in educational practice. This paper introduces a novel framework, \textit{Cognitive Structure Generation} (CSG), in which we first pretrain a \textit{Cognitive Structure Diffusion Probabilistic Model} (CSDPM) to generate students' cognitive structures from educational priors, and then further optimize its generative process as a policy with hierarchical reward signals via reinforcement learning to align with genuine cognitive development levels during students' learning processes. Experimental results on four popular real-world education datasets show that cognitive structures generated by CSG offer more comprehensive and effective representations for student modeling, substantially improving performance on KT and CD tasks while enhancing interpretability.

\end{abstract}

\section{Introduction}
Cognitive structure, originally conceived in topological psychology \cite{lewin2013principles} and later embraced by cognitive psychology in education \cite{piaget1952origins,bruner2009process,ausubel1978educational}, denotes the knowledge system within a student’s mind, manifested as an inherent learning state. Through the learning processes, students continually integrate new concepts and reorganize existing ones to refine their cognitive structures for further learning. Formally, a cognitive structure can be modeled as an evolving \textit{graph} \cite{novak1984learning}, with nodes and edges representing the student’s construction of concepts and inter‑concept relations, respectively \cite{steffe1995constructivism}.

Cognitive structure assessment has long been a central topic in psychometrics \cite{lord2008statistical}. Traditional methods primarily relied on expert‑defined educational priors to directly calculate cognitive structure but lacked sufficient accuracy \cite{tatsuoka2009cognitive,lin2016effects}. Considering that cognitive structure is an inherent learning state, researchers have shifted to indirectly measuring it based on students’ responses to test items. Knowledge tracing (KT) \cite{corbett1994knowledge} and cognitive diagnosis (CD) \cite{leighton2007cognitive} are prototypical tasks. KT predicts the response $r_t$ at time $t$ as \(P_{KT}(r_t) = f_{KT}\bigl(\boldsymbol{h}_t, \boldsymbol{\beta}_t;\Phi\bigr)\), where $\boldsymbol{h}_t$ is the student’s latent state inferred from historical interactions before $t$, $\boldsymbol{\beta}_t$ is the tested item’s features, and $\Phi$ denotes the model parameters \cite{abdelrahman2023knowledge}. CD models the association between response $r$ and student’s cognitive state or ability $\boldsymbol{\theta}$ based on tested item $\beta$ as \(P_{CD}(r) = f_{CD}\bigl(\boldsymbol{\theta}, \boldsymbol{\beta}; \Omega\bigr)\),where $\Omega$ denotes the model parameters \cite{wang2024survey}. Although recently emerged KT \cite{piech2015deep,choi2020towards,zhang2017dynamic} and CD \cite{cheng2019dirt,wang2020neural} models have achieved remarkable performance, they still face two foundational limitations. 

First, both the student’s latent state $\boldsymbol{h}_t$ in KT and the cognitive state or ability $\boldsymbol{\theta}$ in CD are typically narrowed to the student’s construction of individual concept, i.e.\ $\boldsymbol{h}_t,\boldsymbol{\theta}\to\mathbb{R}^L$ (where $L$ is the number of concepts), and thus cannot model the student’s construction of inter‑concept relations necessary for modeling a complete cognitive structure and its holistic evolution during the real learning process. Although some studies have applied graph learning methods on static concept maps \cite{liu2019exploiting,Nakagawa2019Graph,Tong2020Structure} or heterogeneous interaction graphs \cite{DBLP:conf/sigir/Gao0HYBWM0021,DBLP:conf/nips/YangC00ZM024} to obtain enhanced representations of $\boldsymbol{h}_t$ and $\boldsymbol{\theta}$, they only model students’ construction on individual concepts and still do not explicitly model students’ construction of inter‑concept relations. Therefore, our core motivation is to address a long-standing challenge in student modeling: cognitive structure (CS)—the status of the students’ construction of concept and inter-concept relationships \cite{ausubel1978educational}—remains a foundational yet unassessable concept in educational practice.

Second, current KT/CD methods typically intertwine latent state inference with prediction targets, which impairs generalization, particularly in cold-start or uncertain settings, and limits modular reuse. Our intention is not to introduce another KT/CD model. Instead, the key is to propose a novel framework that can decouple cognitive state learning (representation) from outcome prediction (downstream tasks), and produces pretrained cognitive structures that can be seamlessly integrated as plug-and-play inputs into diverse downstream tasks, thereby offering modularity and extensibility.

To bridge this gap, we propose \textbf{Cognitive Structure Generation (CSG)}, a task-agnostic framework that explicitly models CS via generative modeling. Guided by cognitive structure theory \cite{ausubel1978educational} and constructivism \cite{steffe1995constructivism}, CSG aims to learn interpretable, holistic, evolving cognitive structures from their learning interactions that are transferable across tasks. Specifically:

\textbf{First}, consider a cognitive structure is manifested as a graph, we naturally cast \textit{cognitive structure generation} as a \textit{graph generation} task, and propose a \textit{Cognitive Structure Diffusion Probabilistic Model} (CSDPM) for CSG, whose forward diffusion and reverse denoising processes can learn the underlying distribution of real cognitive structures and produce novel ones. However, since real cognitive structures cannot be directly observed for training CSDPM, we devise a rule‐based method to infer students’ construction of concepts and inter‑concept relations from their interaction logs, yielding a set of simulated cognitive structures that inherently incorporate educational priors, which is then used to pretrain the CSDPM and initialize its basic capability for CSG. 

\textbf{Second}, although the cognitive structures sampled from the pretrained CSDPM match the distribution over simulated cognitive structures with educational priors, they are insufficient to reflect the genuine levels of cognitive development \cite{flavell1977cognitive,keil1992concepts} that students achieve through their learning processes. To fill this gap, inspired by \textit{the Structure of the Observed Learning Outcome (SOLO) taxonomy} \cite{biggs2014evaluating} that characterizes five levels of cognitive development (i.e., prestructural, unistructural, multistructural, relational, and extended abstract), we define a fine-grained, hierarchical reward function inspired by the SOLO taxonomy \citep{biggs2014evaluating}, which evaluates generated cognitive structures based on their alignment with observed student interactions. Using these reward signals, we optimize the policy of the denoising process via reinforcement learning to better align generation with cognitive development stages observed in student learning.

To the end, the CSDPM for CSG, pretrained with educational priors and fine-tuned via policy optimization based on SOLO-based reward signals, has been fully trained for cognitive structure generation. These generated cognitive structures can be leveraged for a variety of downstream student‐modeling tasks in educational domain. To the best of our knowledge, we are the \textbf{first} to: 
\begin{itemize}
    \item Reformulate cognitive structure modeling as a cognitive structure generation task;
    \item Apply Graph Diffusion Probabilistic Models for CSG;
    \item Introduce an end-to-end CSG framework with a two-stage pretraining–finetuning procedure;
    \item Guided by educational theories, propose a rule-based educational prior method for simulating cognitive structures and a SOLO-based hierarchical objectives for policy optimization.
\end{itemize}
Experimental results on four popular real-world education datasets show that cognitive structures generated by CSG offer more comprehensive and effective representations for student modeling, substantially improving performance on KT and CD tasks while enhancing interpretability.


\section{Related Works}

\textbf{Cognitive Structure Modeling}. The students’ cognitive structures \cite{lewin2013principles,piaget1952origins,bruner2009process,ausubel1978educational} represent their internal knowledge system, an evolving graph whose nodes reflect their construction of concepts and whose edges capture their construction of inter‐concept relations \cite{novak1984learning,steffe1995constructivism}. Traditional psychometric approaches derive such structures from expert‐defined rules, which limits personalization and accuracy \cite{lord2008statistical,tatsuoka2009cognitive,lin2016effects}. Considering that cognitive structure is an inherent learning state, researchers have shifted to indirectly measuring it based on students’ responses to test items, e.g., knowledge tracing (KT) and cognitive diagnosis (CD). 

From the KT perspective \cite{piech2015deep,choi2020towards,zhang2017dynamic}, cognitive structures are implicitly approximated via students’ learning states (also termed hidden states or knowledge states) inferred from response logs. This includes theory‐guided state models \cite{DBLP:conf/kdd/GuLDLZ25,DBLP:conf/www/SunYWLLS24}, mastery pattern classifiers \cite{briggs2017challenges,cui2016statistical}, and encoder–decoder architectures \cite{DBLP:conf/www/LiLWLHYZS24,DBLP:conf/www/LiuSQZ24,DBLP:conf/nips/ChenWLCZHW23}. Some KT methods enrich these states with static concept maps or heterogeneous interaction graphs \cite{liu2019exploiting,Nakagawa2019Graph,Tong2020Structure,DBLP:conf/sigir/Gao0HYBWM0021,DBLP:conf/nips/YangC00ZM024}, yet they typically emphasize concept mastery without modeling the formation of inter‐concept relations. From the CD perspective \cite{leighton2007cognitive,cheng2019dirt,wang2020neural}, models aim to identify fine‐grained cognitive attributes or abilities underlying observed responses. While some approaches introduce additional features \cite{DBLP:journals/tkde/LiuHYCXSH21,DBLP:conf/kdd/XuHLSLCW023,DBLP:conf/kdd/Zhou0WWHT0CM21}, address data distribution issues \cite{DBLP:conf/aaai/ChengDLNTXN25,DBLP:conf/icdm/Zhang0SMZ0Z23}, or optimize network structures \cite{DBLP:conf/nips/YangYTYM023,yang2023evolutionary}, they also tend to focus on the correctness of individual concepts, overlooking the holistic evolution of cognitive structures. 

A recent attempt \cite{chen2024enhancingcognitivediagnosismodeling} to model cognitive structure state still relies on a predefined concept graph and treats node and edge construction independently, failing to capture their coupled dynamics. To our knowledge, we are the first to explicitly formulate the task of cognitive structure generation and present a unified framework for its holistic modeling.

\textbf{Graph Diffusion Probabilistic Models}. Graph generation has long relied on traditional deep generative frameworks (e.g., auto‐regressive models \cite{DBLP:conf/nips/LiaoLSWHDUZ19}, VAEs \cite{DBLP:conf/nips/LiuABG18}, GANs \cite{DBLP:conf/icml/MartinkusLPW22}, and normalizing flows \cite{luo2021graphdf}) to capture complex graph distributions. More recently, diffusion probabilistic models (DPMs) \cite{ho2020denoising} have emerged as a powerful new trend for graph generation \cite{zhang2023survey}. Continuous‐time graph DPMs (e.g., EDP‐GNN \cite{niu2020permutation}, GDSS\cite{jo2022score}, DruM \cite{jo2023graph}) learn to denoise Gaussian‐corrupted graph representations \cite{song2020score} but can struggle to preserve graph sparsity. To address this, discrete diffusion methods like DiGress \cite{DBLP:conf/iclr/VignacKSWCF23} replace continuous noise with categorical transitions, achieving strong results on complex benchmarks. To our knowledge, we are the first to introduce a graph diffusion probabilistic models for CSG and learn CSDPM with educational priors.

\textbf{Optimizaion of DPMs}. Reinforcement learning (RL) has been widely used to steer graph generators toward downstream objectives. Traditional methods \cite{DBLP:conf/nips/SuttonMSM99,DBLP:journals/corr/abs-1810-08678} rely on custom environments and exhibit high computational cost. Diffusion models (DPMs) have been aligned to external rewards in vision: DPOK \cite{DBLP:conf/nips/FanWDLRBAG0L23} and DDPO \cite{DBLP:conf/iclr/BlackJDKL24} treat the reverse diffusion as a Markov decision process and apply policy gradients to optimize black‐box reward signals, and DPM alignment has been extended to graphs by GDPO \cite{DBLP:conf/nips/LiuDPLL024}, which introduces a eager policy gradient. Thus, we propose a SOLO-based rewards to optimize the CSDPM, which is effectiveness for aligning with cognitive development levels.

\section{The CSG Framework}

\subsection{Problem Formulation}
Suppose that in a learning system \(\mathcal{L}=\langle S, Q, K, R\rangle\), there are $N$ students, $M$ questions, and $L$ knowledge concepts, denoted as $S = \{s_{i}\}_{i=1}^N$, $Q = \{q_{j}\}_{j=1}^M$, and $K = \{k_{l}\}_{l=1}^L$, respectively. Students practice questions from the pool $Q$, generating response logs \(R = \{r_{ij}|\text{student }s_i\text{ answered questions }q_j\}\), where $r_{ij} = 1$ if student $s_i$ answered question $q_j$ correctly, and $r_{ij} = 0$ otherwise. For each student $s_i$, the sequence of historical interactions up to timestamp $T$ can be represented as \(X_i^T = \{(q_{j}, r_{ij})^t\}_{t=1}^T\), where \(j\) indexes over \(Q\) and \(q_{j} \in Q\). 


A student $s_i$'s cognitive structure at timestamp $T$ can be defined as a graph: \(\mathcal{G}_i^T = (\mathcal{V}_i^T, \mathcal{E}_i^T)\), where nodes $\mathcal{V}_i^T \in \mathbb{R}^{L\times c}$ represent $s_i$'s construction of concepts in $K$, and edges $\mathcal{E}_i^T \in \mathbb{R}^{L\times L \times c}$ represent $s_i$'s construction of inter-concept relations, where \(c\) is the number of construction state space, which denotes whether a concept or relation is ``constructed'' or ``unconstructed'', forming a discrete state space of size. Since we treat the cognitive structure as an undirected graph, all subsequent operations are applied exclusively to the upper-triangular entries $\mathcal{E}^+$ of $\mathcal{E}$, after which the matrix is symmetrized. Our goal is to generate $s_i$'s cognitive structure $\mathcal{G}_i^T$ based on $X_i^T$. Formally, a function \(f_{CSG}\) is used to learn a mapping for CSG, \( f_{CSG}: X_i^T \rightarrow \mathcal{G}_i^T \).

In this work, we propose a \textit{Cognitive Structure Diffusion Probabilistic Model} (CSDPM) to implement the mapping function $f_{CSG}$. Specifically, the proposed CSDPM first learns to generate cognitive structures via pretraining on educational priors, and is then fine-tuned through policy optimization guided by educational objectives. The holistic cognitive structures sampled from the optimized CSDPM can be used to improve the performance of downstream student modeling tasks (e.g., KT and CD): $P_{KT}(r_{ij}^{T+1}) = f_{KT}\bigl(\mathcal{G}_i^T, \boldsymbol{\beta}(q_{j}^{T+1});\Phi\bigr)$ and $P_{CD}(r_{ij}) = f_{CD}\bigl(\mathcal{G}_i^T, \boldsymbol{\beta}(q_{j}); \Omega\bigr)$, respectively.

\subsection{Framework Overview}
\begin{figure}
        \centering
        \includegraphics[width=\linewidth]{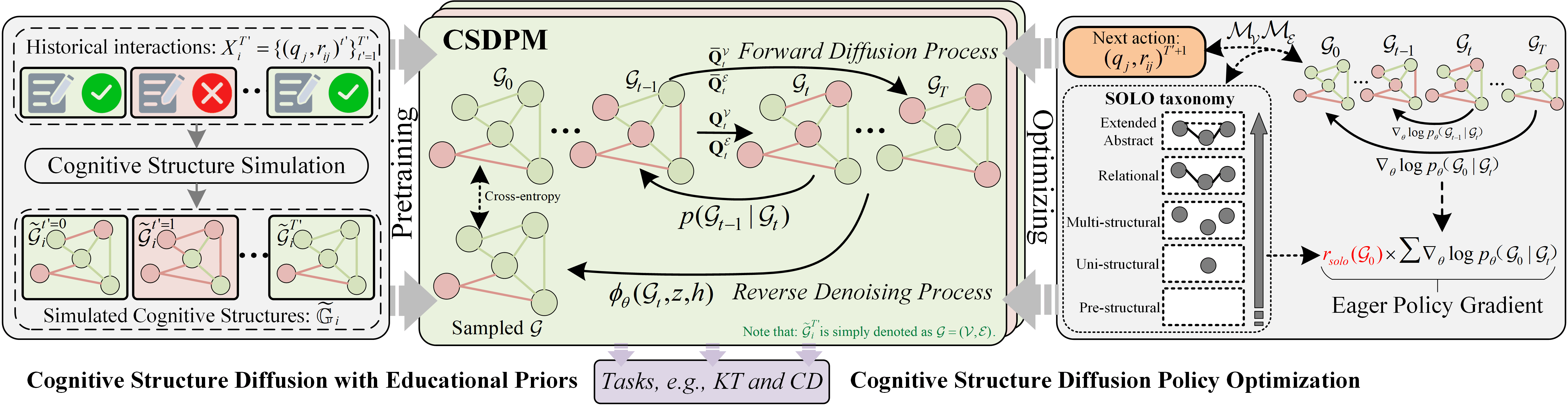}
        \caption{\textbf{(Overview).} The CSG includes two stages: Cognitive Structure Diffusion with Educational Priors and Cognitive Structure Diffusion Policy Optimization. In first stage, the Cognitive Structure Simulation module (left) produces simulated cognitive structures that are used to pretrain the CSDPM. The second stage introduces a SOLO‐based reward to optimize the CSDPM’s policy via RL (right). Once pretrained and optimized, the CSDPM can generates cognitive structures, whose effectiveness is validated on KT and CD tasks through response prediction.
        }
        \label{fig:csg}
\end{figure}

The overall architecture of CSG is illustrated in Fig.\ref{fig:csg}. 

\subsection{Cognitive Structure Diffusion with Educational Priors}
This section presents the cognitive structure diffusion probabilistic model (CSDPM) pretrained with educational priors, whose reverse denoising processes can learn the underlying distribution of real cognitive structures and produce novel ones. Unlike other graph generation domains \cite{DBLP:conf/nips/LiuXLJ24,DBLP:conf/nips/ZhangLWFL24,DBLP:conf/icml/TrivediRA0DKLPA24,DBLP:conf/nips/ZhaoDDKT21}, training a cognitive structure diffusion model ideally requires access to ground-truth cognitive structures, which are not available in practice. To approximate this, we incorporate an educational prior grounded in well-established theories of cognitive structure \cite{ausubel1978educational} and constructivist learning \cite{steffe1995constructivism}. Specifically, we use a simple rule-based generation process to simulate how students may construct and connect concepts, which serves as a proxy for pretraining.

\textbf{Cognitive Structure Simulation}. To simulate student $s_i$’s cognitive structure \(\tilde{\mathcal{G}}_i^T\) for pretraining CSDPM, inspired by \cite{DBLP:journals/bjet/LinCLC16}, we propose a rule-based method to calculate the constr\underline{u}ction state $\tilde{v}_{il}^T\in \mathcal{V}_i^T$ \underline{o}f \underline{c}oncept $k_l$ by function \(f_{UOC}\), the constr\underline{u}ction state $\tilde{e}_{i,a\!-\!b}^T\in {\mathcal{E}_i^T}^+$ \underline{o}f \underline{r}elation between concepts $k_a$ and $k_b$ by function \(f_{UOR}\) based on historical interactions \(X_i^T\):
\begin{equation}\label{eq:uoc}
    \tilde{v}_{il}^T:f_{UOC}(k_l,X_i^T) = \frac{\sum\nolimits_{(q_{j}, r_{ij})^t\in X_i^T}(\omega_{l,j}\cdot \frac{r_{ij}}{|T|})}{\sum\nolimits_{(q_{j}, r_{ij})^t\in X_i^T}(\omega_{l,j}\cdot \frac{1}{|T|})},
\end{equation}
\begin{equation}\label{eq:uor}
    \tilde{e}_{i,a\!-\!b}^T:f_{UOR}(k_a,k_b,X_i^T) = \frac{\sum\nolimits_{(q_{j}, r_{ij})^t\in X_i^T}\Bigl[\mathbf{1}\{\omega_{a,j}>0\land\omega_{b,j}>0\}\,(\omega_{a,j}+\omega_{b,j})\frac{r_{ij}}{|T|}\Bigr]}{\sum\nolimits_{(q_{j}, r_{ij})^t\in X_i^T}\Bigl[\mathbf{1}\{\omega_{a,j}>0\land\omega_{b,j}>0\}\,(\omega_{a,j}+\omega_{b,j})\frac{1}{|T|}\Bigr]},
\end{equation}
where \(\tilde{v}_{il}^T,\tilde{e}_{i,a\!-\!b}^T \in \Delta^c\) denote probability distributions over the \(c\) construction state space while could map into the discrete construction categories. \(\omega_{l,j}\) represent the predefined weight of concept $k_l$ tested in question $q_j$, as computed by the procedure in \cite{DBLP:journals/kbs/YangQLGRZW22}. \(|T|\) is the length of the interaction sequence and serves for normalization. $\mathbf{1}\{\omega_{a,j}>0\land\omega_{b,j}>0\}$ is the indicator function that equals \(1\) when both $\omega_{a,j}$ and $\omega_{b,j}$ are positive, and \(0\) otherwise. Suppose a student completes five “Law of Sines” questions, each equally weighted, and two of these also involve the “Law of Cosines.” If the student answers two of the “Law of Sines” questions correctly and one of the joint “Sine–Cosine Theorem” questions correctly, then \(v^T(\text{Sine Theorem}) = 0.625, e^T(\text{Sine–Cosine Theorem}) = 0.5\). Please see the attachment for a detailed example of the calculation process in Appendix \ref{app:cal}.


Thus, for any student $s_i$ at any timestamp $T$, the simulated cognitive structure $\tilde{\mathcal{G}}_i^T\in\tilde{\mathbb{G}}$ can be obtained by the above method, where $\tilde{\mathbb{G}}$ denotes the simulated dataset used to pretrain the CSDPM through the forward diffusion and reverse denoising processes, as detailed below. 

Note that we omit the student subscript $i$ and superscript $T$ when unambiguous, writing $\tilde{\mathcal G}_i^T$, $\tilde v_{il}^T$, $\tilde e_{i,a\!-\!b}^T$ simply as $\mathcal G$, $v$, and $e$. To distinguish between interaction timestamps and diffusion steps, we denote the former by $T'$ and reserve $T$ for the latter if necessary.

\textbf{Forward Diffusion Process}. Our CSDPM involves a forward diffusion process $q(\mathcal{G}_{1:T}|\mathcal{G}_{0}) = \prod_{t=1}^{T} q(\mathcal{G}_{t}|\mathcal{G}_{t-1})$, which gradually corrupts the initial simulated cognitive structure distribution $q(\mathcal{G}_{0})$ into a simple noise distribution $q(\mathcal{G}_{T})$ after $T$ diffusion steps. The transition distribution \(q(\mathcal{G}_{t}|\mathcal{G}_{t-1})\) can be factorized into transitions of construction state space for all nodes and edges: 
\begin{equation}
    q(\mathcal V_{t}|\mathcal V_{t-1})=\prod\nolimits_{v \in \mathcal V}q({v}_{t}|{v}_{t-1}), \quad q({\mathcal E}_{t}|{\mathcal E}_{t-1})=\prod\nolimits_{e \in \mathcal E^{+}}q({e}_{t}|{e}_{t-1}).
\end{equation}
For each $v$, we define \(q({v}_{t}|{v}_{t-1}) = Cat({v}_{t}; {v}_{t-1}\boldsymbol{Q}_{t}^{v})\), where \(\boldsymbol{Q}_{t}^{v} = \alpha_{t}\boldsymbol{I} + (1 - \alpha_{t})\frac{\mathbf{1}_c\mathbf{1}_c^{\top}}{c}\), and $\alpha_{t}$ decays from 1 to 0 as $t$ increases \cite{DBLP:conf/nips/AustinJHTB21}. The term \(\frac{\mathbf{1}_c\mathbf{1}_c^{\top}}{c}\) represents a \(c \times c\) matrix where every entry is \(1/c\), corresponding to a uniform distribution over all possible states (i.e., maximally random noise). When \(\alpha_t = 1\), we have \(\boldsymbol{Q}_{t}^{v} = \boldsymbol{I}\), meaning no noise is added and the signal (graph) remains intact. For intermediate steps \(t\), \(\boldsymbol{Q}_{t}^{v}\) is a linear combination of the original signal and uniform noise, where larger values of \(\alpha_{t}\) inject less noise, and smaller values inject more.
The posterior distributions admit closed-form using Bayes rule:
\begin{equation}
    q(v_{t}|v_{0}) = \mathrm{Cat}\bigl(v_{t};\,v_{0}\,\bar{\boldsymbol{Q}}_{t}^{v}\bigr), \quad q(v_{t-1}|v_t,v_0) = \text{Cat}\left(v_{t-1}; \frac{v_t \boldsymbol{Q}_t^\top \odot v_0 \bar{\boldsymbol{Q}}_{t-1}}{v_0 \bar{\boldsymbol{Q}}_t v_t^\top}\right),
\end{equation}
where $\bar{\boldsymbol{Q}}_{t}^{v}=\boldsymbol{Q}_{1}^{v}\boldsymbol{Q}_{2}^{v}\cdots \boldsymbol{Q}_{t}^{v}$, \(\mathrm{Cat}(;)\) denotes a categorical distribution over \(c\) construction state space, and $\odot$ is element-wise production. The forward diffusion process drives each node toward the uniform distribution \(q({v}_{T}|{v}_{0}) \approx Cat({v}_{T}; \frac{\mathbf{1}_c}{c})\) when \(\lim_{t \to \infty} \alpha^t = 0\). The transition distribution for edges is defined analogously and omitted for brevity. 

\textbf{Reverse Denoising Process}. Given the forward diffusion process, a parametric reverse denoising process \(p_{\theta}(\mathcal{G}_{0:T}) = p(\mathcal{G}_{T})\prod\nolimits_{t=1}^{T}p_{\theta}(\mathcal{G}_{t-1}|\mathcal{G}_{t})\) is learned to recover the cognitive structure distribution from the approximated uniform noise distribution $p(\mathcal{G}_{T})\approx q(\mathcal{G}_{T})$. Specifically, we model the reverse transition distribution as a product of categorical distributions of nodes and edges:
\begin{equation}
    p_{\theta}(\mathcal{G}_{t-1}|\mathcal{G}_{t}) = \prod\nolimits_{v \in \mathcal V} p_{\theta}({v}_{t-1}|\mathcal{G}_{t}) \prod\nolimits_{e \in \mathcal E^{+}} p_{\theta}({e}_{t-1}|\mathcal{G}_{t}).
\end{equation}
In alignment with the standard $x_0$-parameterization \cite{hasselt2010double,DBLP:conf/nips/KarrasAAL22} used in diffusion models, each node and edge conditional distribution can be formulated as follows:
\begin{equation}
    p_{\theta}(v_{t-1}|\mathcal{G}_t)=\sum_{v_0\in\Delta^c}q\bigl(v_{t-1}|v_t,\,v_0\bigr)p_{\theta}\bigl(v_0|\mathcal{G}_t\bigr), \quad
    p_{\theta}(e_{t-1}|\mathcal{G}_t)=\sum_{e_0\in\Delta^c}q\bigl(e_{t-1}|e_t,\,e_0\bigr)p_{\theta}\bigl(e_0|\mathcal{G}_t\bigr),
\end{equation}
where a neural network $p_{\theta}$ predicts the posterior distributions $p_{\theta}\bigl(v_0|\mathcal{G}_t\bigr)$ and $p_{\theta}\bigl(e_0|\mathcal{G}_t\bigr)$ given the noisy cognitive structure $\mathcal{G}_{t}$ and $\Delta^c$ is the categorical construction state space.

\textbf{Training Objective}. The CSDPM is trained on the simulated cognitive structure dataset $\tilde{\mathbb{G}}$ by maximizing the following objective:
\begin{equation}
    J_{\text{CSDPM}}(\theta) = \mathbb{E}_{\mathcal{G}_{0}\sim \tilde{\mathbb{G}}, t\sim \mathcal U\llbracket 1,T \rrbracket} \left[ \mathbb{E}_{q(\mathcal{G}_{t}|\mathcal{G}_{0})}\left[\log p_{\theta}(\mathcal{G}_{0}|\mathcal{G}_{t}) \right] \right],
\end{equation}
where $t$ is uniformly sampled from $\llbracket 1,T \rrbracket$. After pretraining, cognitive structures with educational priors can be generated by first sampling $\mathcal{G}_{T}$ from $p(\mathcal{G}_{T})$ and subsequently iteratively sampling $\mathcal{G}_{t-1}$ from $p_{\theta}(\mathcal{G}_{t-1}|\mathcal{G}_{t})$, yielding a trajectory $(\mathcal{G}_{T}, \mathcal{G}_{T-1}, \dots, \mathcal{G}_{0})$ for prior cognitive structure generation.

\textbf{Parametrization}. Building on the extended Graph Transformer network \cite{DBLP:journals/corr/abs-2012-09699,DBLP:conf/iclr/VignacKSWCF23}, we parameterize the reverse‐denoising distribution $p_{\theta}$, which takes a noisy cognitive structure $\mathcal{G}_{t}=(\mathcal{V}_{t},\mathcal{E}_{t})$ as input and outputs a predicted distribution over the clean cognitive structure. We retain the graph‐theoretic feature integration in \cite{DBLP:conf/iclr/VignacKSWCF23}, and additionally embed the current sample’s historical interaction sequence $X^{T'}$ to serve as guidance features. The algorithm is summarized in Appendix \ref{sec:pretrain-csdpm}.

\subsection{Cognitive Structure Diffusion Policy Optimization}
We define a fine-grained, hierarchical reward function inspired by the SOLO taxonomy \citep{biggs2014evaluating}, which evaluates generated cognitive structures based on their alignment with observed student interactions. Using these reward signals, we optimize the policy of the denoising process via reinforcement learning to better align generation with cognitive development stages observed in student learning.


\textbf{Standard Markov Decision Process Formulation}. A Standard MDP is specified by the tuple $(\mathcal S,\mathcal A,\mathcal P,r,\rho_0)$, where $\mathcal S$ is the space of all possible states, $\mathcal A$ is the space of all available actions, $\mathcal P(\boldsymbol{s}'|\boldsymbol{s}, \boldsymbol{a})$ is the transition function, $r(\boldsymbol{s},\boldsymbol{a})$ is the reward signal, and $\rho_0$ is the initial‐state distribution. Under a parameterized policy $\pi_\theta(\boldsymbol{a}|\boldsymbol{s})$, an agent generates a trajectory $\boldsymbol{\tau}=(\boldsymbol{s}_0,\boldsymbol{a}_0,\boldsymbol{s}_1,\boldsymbol{a}_1,\dots,\boldsymbol{s}_T,\boldsymbol{a}_T)$ by sampling $\boldsymbol{s}_0\sim\rho_0$, then repeatedly selecting $\boldsymbol{a}_t\sim\pi_\theta(\,\cdot|\boldsymbol{s}_t)$, receiving reward score $r(\boldsymbol{s}_t,\boldsymbol{a}_t)$, and transitioning via $\boldsymbol{s}_{t+1}\sim P(\,\cdot|\boldsymbol{s}_t,\boldsymbol{a}_t)$. The symbol \(\sim\) denotes sampling from a distribution, consistent with standard usage in probabilistic modeling. The cumulative reward return of $\boldsymbol{\tau}$ is $\mathcal R(\boldsymbol{\tau})=\sum_{t=0}^T r(\boldsymbol{s}_t,\boldsymbol{a}_t)$, and the learning objective is to maximize its expectation \(\mathcal J_{\mathrm{RL}}(\theta)=\mathbb{E}_{\boldsymbol{\tau}\sim p(\boldsymbol{\tau}|\pi_\theta)}\bigl[\mathcal R(\boldsymbol{\tau})\bigr]\). By the policy‐gradient theorem \cite{grondman2012survey}, \(\mathcal J_{\mathrm{RL}}(\theta)\) can be optimized with policy gradient \(\nabla_\theta \mathcal J_{\mathrm{RL}}(\theta)\), as follows:
\begin{equation}
    \nabla_\theta \mathcal J_{\mathrm{RL}}(\theta)
    =\mathbb{E}_{\boldsymbol{\tau}\sim p(\boldsymbol{\tau}|\pi_\theta)}\Bigl[\sum\nolimits_{t=0}^T\nabla_\theta\log\pi_\theta(\boldsymbol{a}_t| \boldsymbol{s}_t)\,\mathcal R(\boldsymbol{\tau})\Bigr],
\end{equation}
which can be estimated via Monte‐Carlo sampling using the REINFORCE algorithm \cite{sutton1998reinforcement}.

\textbf{Mapping \(T\)-steps Reverse Denoising Process of CSDPM to \(T\)-steps MDP}. The pretrained CSDPM indeed defines a distribution of sampled cognitive structure \(p_{\theta}(\mathcal{G}_{0})\) through its reverse denoising process \(p_{\theta}(\mathcal{G}_{0:T})\). In principle, we would like to maximize the expected reward \(\mathcal J_{\mathcal G_0}(\theta)=\mathbb{E}_{\mathcal{G}_{0}\sim p_{\theta}(\mathcal{G}_{0})}\bigl[r(\mathcal{G}_{0})\bigr]\). However, although pre-training has taught the CSDPM how to sample from each step $p_\theta(\mathcal G_{t-1}|\mathcal G_t)$, it cannot yield a tractable, closed-form expression for the marginal likelihood $p_\theta(\mathcal{G}_{0})$ \cite{ho2020denoising}. Moreover, the reward signal $r(\cdot)$ is treated as a black-box function of the sampled cognitive structure, so its gradient $\frac{\partial r(\mathcal G_0)}{\partial \mathcal G_0}$ is unavailable \cite{DBLP:conf/iclr/BlackJDKL24}. In order to optimize the reverse denoising process of CSDPM for specific reward signals, inspired by \cite{DBLP:conf/nips/FanWDLRBAG0L23,DBLP:conf/nips/LiuDPLL024}, we formulate it as a \(T\)-step MDP, as follows:
\begin{align}
\begin{split}
    &\boldsymbol{s}_t \triangleq (\mathcal{G}_{T-t},T-t), \quad \boldsymbol{a}_t \triangleq \mathcal{G}_{T-t-1}, \\
    &\pi_\theta(\boldsymbol{a}_t|\boldsymbol{s}_t) \triangleq p_\theta(\mathcal G_{T-t-1}|\mathcal G_{T-t},T-t), \quad \mathcal P(\boldsymbol{s}_{t+1}|\boldsymbol{s}_t, \boldsymbol{a}_t) \triangleq (\delta_{\mathcal G_{T-t-1}},\delta_{T-t-1}),\\
    &r(\boldsymbol{s}_t, \boldsymbol{a}_t) \triangleq r(\mathcal{G}_{0}) \text{ if } t=T, \quad r(\boldsymbol{s}_t, \boldsymbol{a}_t) \triangleq 0 \text{ if } t<T,
\end{split}
\end{align}
where the initial state \(\boldsymbol{s}_0\), the final state \(\boldsymbol{s}_T\) correspond to the full the fully noised cognitive structure \(\mathcal G_{T}\), the fully denoised cognitive structure \(\mathcal G_{0}\), respectively. 

\textbf{SOLO-based Hierarchical Reward Function}. After formulating the reverse denoising process of CSDPM as a MDP, we can optimize it for specific reward signals, which should ideally reflect the genuine levels of cognitive development that students achieve through their learning processes. Inspired by the SOLO taxonomy \cite{biggs2014evaluating}, we propose a fine-grained, hierarchical reward function that scores the generated cognitive structures according to the five levels: prestructural, unistructural, multistructural, relational, and extended abstract, which correspond to progressively better construction of concepts and inter-concept relations within more sophisticated cognitive structure. 

Specifically, given a student \(s_i\)'s sampled \(\mathcal{G}_{0} = (\mathcal V_0,\mathcal E_0)\) and the real interaction \((q_{j}, r_{ij})^{T'+1}\) at next interaction, the matching degree is calculated based on logical equivalence between the student's real response \(r_{ij}\) with student \(s_i\)'s constructions \(\mathcal V_{q_j} \in \mathcal{V}_{0}\) of the tested concepts and construction \(\mathcal E_{q_j} \in \mathcal{E}_{0}\) of the tested inter-concept relations. The matching functions \(\mathcal M_{\mathcal V}\) and \(\mathcal M_{\mathcal E}\) are as follows:
\begin{equation}
    \mathcal M_{\mathcal V}=\frac{1}{|\mathcal V_{q_j}|} \sum\nolimits_{v\in \mathcal V_{q_j}}(r_{ij}\veebar v), \quad \mathcal M_{\mathcal E}=\frac{1}{|\mathcal E_{q_j}|} \sum\nolimits_{e\in \mathcal E_{q_j}}(r_{ij}\veebar e),
\end{equation}
where \(\veebar\) is the XNOR operation. Then, the SOLO-based reward function $r_{solo}(\mathcal G_0)$ is as follows:
\begin{equation}
    r_{solo}(\mathcal G_0) \;=\;
\begin{cases}
r_{1}, & \mathcal M_{\mathcal V}=0 , \\
r_{2}, & 0 < \mathcal M_{\mathcal V} < 0.5,\\
r_{3}, & \mathcal M_{\mathcal V} \geq 0.5 \land \mathcal M_{\mathcal E} < 0.5,\\
r_{4}, & 0.5 \leq \mathcal M_{\mathcal V} < 1 \land 0.5 \leq \mathcal M_{\mathcal E} < 1,\\
r_{5}, & (\mathcal M_{\mathcal V}=1 \land \mathcal M_{\mathcal E}\geq0.5) \lor (\mathcal M_{\mathcal V}\geq0.5 \land \mathcal M_{\mathcal E}=1),
\end{cases}
\end{equation}
where $r_1<r_2<r_3<r_4<r_5$ are increasing constants reflecting the five SOLO levels from lowest (prestructural) to highest (extended abstract). The matching degrees \(\mathcal M_{\mathcal V}\) and \(\mathcal M_{\mathcal E}\) measure how well the generated cognitive structure aligns with the student’s responses at the concept and inter-concept relation levels, respectively. These two quantities are inspired by the five SOLO taxonomy levels, which characterize progressively deeper levels of cognitive alignment:
\begin{itemize}
    \item \textit{Pre-structural}: No meaningful concept alignment
    \item \textit{Uni-structural}: Alignment of a single or few concepts
    \item \textit{Multi-structural:} Alignment of multiple concepts, few relations
    \item \textit{Relational}: Alignment of multiple concepts and multiple relations
    \item \textit{Extended abstract}: Alignment of almost all concepts and relations
\end{itemize}

Since \(\mathcal M_{\mathcal V}\) and \(\mathcal M_{\mathcal E}\) are both normalized to lie in the range \([0,1]\), we adopt 0.5 as a natural threshold to distinguish between ``few'' and ``multiple'' alignment. For example, when \(0 < \mathcal M_{\mathcal V} < 0.5\), it corresponds to partial concept alignment which maps to the uni-structural level and is thus assigned reward \(r_2\). We will evaluate the performance of CSDPM for CSG under various reward configurations in the experimental section.


\textbf{Policy Gradient Estimation.} Based on the MDP as reverse denoising process of CSDPM, an agent that acts in this MDP can produce the CSG trajectory $ \boldsymbol{\tau}=(\mathcal{G}_{T}, \mathcal{G}_{T-1}, \dots, \mathcal{G}_{0})$, where \(\boldsymbol{\tau} \sim p(\boldsymbol{\tau}|\pi_\theta) = p_{\theta}(\mathcal{G}_{0:T})\). The cumulative reward return of \(\boldsymbol{\tau}\) is \(\mathcal {R}(\boldsymbol{\tau})=\sum_{t=0}^T r(\boldsymbol{s}_t,\boldsymbol{a}_t) = r_{solo}(\mathcal{G}_{0})\) and  the learning objective is to maximize its expectation \(\mathcal J_{\mathrm{RL}}(\theta)=\mathbb{E}_{\boldsymbol{\tau}\sim p(\boldsymbol{\tau}|\pi_\theta)}\bigl[\mathcal R(\boldsymbol{\tau})\bigr] = \mathbb{E}_{\boldsymbol{\tau}\sim p_{\theta}(\mathcal{G}_{0:T})}\bigl[r_{solo}(\mathcal{G}_{0})]\) that is equivalent to \(\mathcal J_{\mathcal G_0}(\theta)\). Similarly, \(\mathcal J_{\mathrm{RL}}(\theta)\) can be optimized with policy gradient \(\nabla_\theta \mathcal J_{\mathrm{RL}}(\theta)\), as follows:
\begin{equation}
    \nabla_\theta J_{\mathrm{RL}}(\theta)
=\mathbb E_{\tau\sim p_{\theta}(\mathcal{G}_{0:T})}\Bigl[r_{solo}(\mathcal G_0)\sum_{t=1}^{T}\nabla_\theta\log p_\theta\bigl(\mathcal G_{t-1}|\mathcal G_{t}\bigr)\Bigr].
\end{equation}
To alleviate the high variance and instability of standard REINFORCE on discrete graph diffusion, following \citet{DBLP:conf/nips/LiuDPLL024}, we employ the \textit{eager} policy gradient term $\nabla_\theta \log p_\theta(\mathcal G_{0}| \mathcal G_t)$, which directly reinforces the likelihood of fully denoised, high-reward end states (i.e., the clean cognitive structures after \(T\) reverse denoising steps), rather than distributing credit iteratively via the term $\nabla_\theta \log p_\theta(\mathcal G_{t-1}| \mathcal G_t)$. With Monte Carlo estimation, the policy gradient can be modified as follows:
\begin{equation}
 \nabla_\theta J_{\mathrm{RL}}(\theta) \approx \frac{1}{|\mathcal D|} \sum_{d=1}^{|\mathcal D|} \;\frac{T}{\mathcal T_{d}}
\sum_{t\in \mathcal T_{d}}
r_{solo}\bigl(\mathcal G^{(d)}_{0}\bigr)\;\nabla_{\theta}\log p_{\theta}\bigl(\mathcal G^{(d)}_{0}|\mathcal G^{(d)}_{t}\bigr)\,,
\end{equation}
where $\mathcal D$ is the set of sampled trajectories, $\mathcal T_{d}\subseteq \llbracket 1,T \rrbracket$ is the random subset of timesteps chosen for the $d$-th trajectory. The estimator treats all $\mathcal D$ trajectories ending in the same \(\mathcal G_0\) as an equivalence class and optimizes over these classes. The algorithm is summarized in Appendix \ref{sec:csdpm-policy-opt}.

To this end, we introduce the two-stage process for CSG. For the pretraining stage, the simulated cognitive structures are designed as a pragmatic way to bootstrap the model in the absence of targets. These structures are derived from well-established educational principles, and aim to approximate common developmental patterns such as the accumulation and gradual integration of concepts. Although simplified, they provide the model with a reasonable prior over what valid cognitive structures might look like, rather than forcing it to learn from pure noise. For the policy optimization stage, we adopt a SOLO-taxonomy–based hierarchical reward function, which evaluates the quality of generated structures by their alignment with increasingly sophisticated levels of cognitive development. This stage allows the model to refine the pretrained representations and move beyond the handcrafted assumptions imposed during pretraining. 

The overall design is inspired by LLM training: a general pretraining stage using rule-based supervision to learn an initial representation, followed by targeted reinforcement learning to better align with desired properties / rewards. This two-stage process enables the model to overcome the lack of explicit ground truth while progressively reducing reliance on heuristic assumptions.



\section{Experiments}\label{sec:experiments}

\subsection{Downstream Models for CSG}
Since there is no ground truth cognitive structures available, we assess the effectiveness of our generated cognitive structures by using its representation on downstream tasks. we take two popular student modeling tasks i.e., knowledge tracing (KT) and cognitive diagnosis (CD), and implement two models, CSG-KT and CSG-CD. Specifically, we first employ an \textit{edge-aware hard-clustering graph pooling} method proposed by \citet{DBLP:journals/corr/abs-2308-11909} to effectively represent students' node and edge construction features together as their cognitive state vectors based on their generated cognitive structures. Then, we concatenate the cognitive state vector with the embedding vector of tested question, and feed them into the output layers of the standard DKT \cite{piech2015deep} and NCD \cite{wang2020neural} to predict students' responses to the tested questions. For CSG-KT, the prediction function is formulated as:
\begin{equation}
    P_{KT}(r_{ij}^{T+1}) = f_{KT,\Phi}:\sigma \left(\text{FC}\left(\text{Pooling}(\mathcal{G}_i^T)\oplus emb(\boldsymbol{\beta}(q_{j}^{T+1}))\right)\right),
\end{equation}
where \(FC\) denotes a fully-connected layer, \(T\) is the timestamp of interaction, and $\sigma$ is the sigmoid activation function. For CSG-CD, the prediction function is defined as:
\begin{equation}
    P_{CD}(r_{ij}) = f_{CD,\Omega}:\sigma \left(\text{FC}\left(\boldsymbol{\mathcal{K}}_j \odot \left((\text{Pooling}(\mathcal{G}_i^T)-\boldsymbol{h}_{diff}) \times \boldsymbol{h}_{disc}\right)\right)\right),
\end{equation}
where $\boldsymbol{\mathcal{K}}_j$ is the vector representing correlations between exercise $q_j$ and its tested concepts $\boldsymbol{\mathcal{K}}_j\in K$. $\boldsymbol{h}_{diff}$ and $\boldsymbol{h}_{disc}$ are attribute feature vectors transformed from $emb(\boldsymbol{\beta}(q_{j}))$. Similar to NCD \cite{wang2020neural}, we apply a fully connected layer with sigmoid activation to the embedding of each question \(q_j\). \(\odot\) is element-wise production. $\sigma$ is the sigmoid activation function. The optimization objective of both CSG-KT and CSG-CD is the cross‐entropy loss between the predicted responses and ground truth.

\subsection{Experimental Settings}

\textbf{Datasets, Baselines and Metrics}.  
We use four real-world datasets of varying scales, i.e., Math1, Math2, and FrcSub\footnote{Math1, Math2 and FrcSub are all available at http://staff.ustc.edu.cn/\%7Eqiliuql/data/math2015.rar}, and NIPS34\footnote{NIPS34 is all available at http://ednet-leaderboard.s3-website-ap-northeast-1.amazonaws.com/}, as shown in Appendix \ref{sec:statisdata}. In order to evaluate the effectiveness of the cognitive structures generated by our CSG, we compare against both classical and state-of-the-art baselines in KT and CD. For KT, we include DKT \cite{piech2015deep}, SAKT \cite{DBLP:conf/edm/PandeyK19}, GKT \cite{Nakagawa2019Graph}, SKT \cite{Tong2020Structure}, GRKT \cite{DBLP:conf/kdd/CuiQJ024}, MIKT \cite{DBLP:conf/www/SunYWLLS24}, ENAS-KT \cite{DBLP:conf/nips/YangYTYM023}, and our CSG-KT. For CD, we consider IRT \cite{cai2016item}, MIRT \cite{ackerman2003using}, NCD \cite{wang2020neural}, RCD \cite{DBLP:conf/sigir/Gao0HYBWM0021}, HyperCDM \cite{DBLP:conf/kdd/ShenQLZJZ24}, DisenGCD \cite{DBLP:conf/nips/YangC00ZM024}, and our CSG-CD. Our goal is to evaluate the usefulness of the generated cognitive structures across downstream tasks, rather than to benchmark a wide range of KT/CD methods. Accordingly, we selected representative baselines from classical (DKT, IRT), structural (GKT, SKT, GRKT, RCD), and sota models (MIKT, ENAS-KT, HyperCDM, DisenGCD) to ensure broad coverage. Similar to the existing methods, we adopt the AUC \cite{bradley1997use}, ACC and RMSE as evaluation metricss. The details of implementaion can be found in Appendix \ref{sec:imp}. 

\begin{table}[t]
\caption{Performance comparison between CSG-KT and KT baselines on different datasets, averaged over five-fold cross-validation. Statistical significance is assessed via the Wilcoxon rank-sum test, with \* ($p<0.05$), \*\* ($p<0.01$), and \*\*\* ($p<0.001$) indicating increasing levels of significance.}
\label{tab:allKT}
\centering
\resizebox{\linewidth}{!}{
\begin{tabular}{c|ccc|ccc|ccc|ccc}
\toprule
\textbf{Datasets}   & \multicolumn{3}{c|}{\textbf{Math1}} & \multicolumn{3}{c|}{\textbf{Math2}} & \multicolumn{3}{c|}{\textbf{FrcSub}}  & \multicolumn{3}{c}{\textbf{NIPS}} \\
\midrule
\textbf{Metrics}    & \textbf{AUC}\(\uparrow\) & \textbf{ACC}\(\uparrow\) & \textbf{RMSE}\(\downarrow\) 
 & \textbf{AUC}\(\uparrow\) & \textbf{ACC}\(\uparrow\) & \textbf{RMSE}\(\downarrow\)
 & \textbf{AUC}\(\uparrow\) & \textbf{ACC}\(\uparrow\) & \textbf{RMSE}\(\downarrow\)
 & \textbf{AUC}\(\uparrow\) & \textbf{ACC}\(\uparrow\) & \textbf{RMSE}\(\downarrow\)\\ 
\midrule
\textbf{DKT}  & 0.7735     & 0.7082    & 0.4524    & 0.7381     & 0.6678    & 0.4600    & 0.8202 & 0.7529     & 0.3392 & 0.6593    & 0.6214    & 0.4690    \\
\textbf{SAKT} & 0.7612     & 0.7017    & 0.4552    & 0.7250     & 0.6583    & 0.4618    & 0.8113 & 0.7513     & 0.3419 & 0.6531    & 0.6176    & 0.4710    \\
\textbf{GKT}  & 0.7843     & 0.7147    & 0.4493    & 0.7463     & 0.6759    & 0.4519    & 0.8247 & 0.7608     & 0.3360 & 0.6841    & 0.6339    & 0.4645    \\
\textbf{SKT}  & 0.7895     & 0.7181    & 0.4489    & 0.7529     & 0.6842    & 0.4492    & 0.8385 & 0.7696     & 0.3338 & 0.6985    & 0.6429    & 0.4637    \\
\textbf{GRKT} & 0.7943     & 0.7242    & 0.4461    & 0.7618     & 0.6976    & 0.4448    & 0.8418 & 0.7754   & 0.3280 & 0.7070    & 0.6501    & 0.4601    \\
\textbf{MIKT} & 0.8030     & 0.7281    & 0.4412    & 0.7701     & 0.7017    & 0.4426    & 0.8472 & 0.7804   & 0.3253 & 0.7147    & 0.6570    & 0.4583    \\
\textbf{ENAS-KT}& 0.8103     & 0.7326    & 0.4334    & 0.7722     & 0.7120    & 0.4405    & 0.8506 & 0.7865  & 0.3207 & 0.7233    & 0.6634    & 0.4565    \\
\midrule
\textbf{CSG-KT} & \textbf{0.8192*} & \textbf{0.7389**} & \textbf{0.4301**} & 
              \textbf{0.7813*} & \textbf{0.7163*} & \textbf{0.4397*} & 
              \textbf{0.8575*} & \textbf{0.7917*} & \textbf{0.3186**} & 
              \textbf{0.7339**} & \textbf{0.6732**} & \textbf{0.4542**} \\
\bottomrule
\end{tabular}}
\end{table}
\begin{table}[t]
\caption{Performance comparison between CSG-CD and CD baselines on different datasets, averaged over five-fold cross-validation. Statistical significance is assessed via the Wilcoxon rank-sum test, with \* ($p<0.05$), \*\* ($p<0.01$), and \*\*\* ($p<0.001$) indicating increasing levels of significance.}
\label{tab:allCD}
\centering
\resizebox{\linewidth}{!}{
\begin{tabular}{c|ccc|ccc|ccc|ccc}
\toprule
\textbf{Datasets}   & \multicolumn{3}{c|}{\textbf{Math1}} & \multicolumn{3}{c|}{\textbf{Math2}} & \multicolumn{3}{c|}{\textbf{FrcSub}}  & \multicolumn{3}{c}{\textbf{NIPS}} \\
\midrule
\textbf{Metrics}    & \textbf{AUC}\(\uparrow\) & \textbf{ACC}\(\uparrow\) & \textbf{RMSE}\(\downarrow\) 
 & \textbf{AUC}\(\uparrow\) & \textbf{ACC}\(\uparrow\) & \textbf{RMSE}\(\downarrow\)
 & \textbf{AUC}\(\uparrow\) & \textbf{ACC}\(\uparrow\) & \textbf{RMSE}\(\downarrow\)
 & \textbf{AUC}\(\uparrow\) & \textbf{ACC}\(\uparrow\) & \textbf{RMSE}\(\downarrow\)\\ 
\midrule
\textbf{IRT}         & 0.7356  & 0.7179 & 0.4279 & 0.7589  & 0.6981 & 0.4516 & 0.7414  & 0.7091  & 0.3944 & 0.7489 & 0.6907 & 0.4516 \\
\textbf{MIRT}        & 0.7482  & 0.7347 & 0.4256 & 0.7699  & 0.7038 & 0.4478 & 0.8086  & 0.7745  & 0.3589 & 0.7589 & 0.7017 & 0.4483 \\
\textbf{NCD}         & 0.7691  & 0.7459 & 0.4084 & 0.7781  & 0.7182 & 0.4456 & 0.8250  & 0.8042  & 0.3498 & 0.7697 & 0.7113 & 0.4412 \\
\textbf{RCD}         & 0.7861  & 0.7584 & 0.4033 & 0.7911  & 0.7275 & 0.4406 & 0.8321  & 0.8178  & 0.3419 & 0.7736 & 0.7171 & 0.4345 \\
\textbf{HyperCDM}    & 0.7876  & 0.7599 & 0.4016 & 0.7972  & 0.7320 & 0.4383 & 0.8417  & 0.8239  & 0.3387 & 0.7821 & 0.7209 & 0.4301 \\
\textbf{DisenGCD}    & 0.7983  & 0.7628 & 0.4001 & 0.8039  & 0.7457 & 0.4324 & 0.8559  & 0.8375  & 0.3342 & 0.7886 & 0.7311 & 0.4275 \\
\midrule
\textbf{CSG-CD} & \textbf{0.8104*}   & \textbf{0.7812**}  & \textbf{0.3933***} & 
               \textbf{0.8193**} & \textbf{0.7608*}   & \textbf{0.4301***} & 
               \textbf{0.8757***} & \textbf{0.8439*}  & \textbf{0.3172***} & 
               \textbf{0.8002*}   & \textbf{0.7499**} & \textbf{0.4251***} \\
\bottomrule
\end{tabular}}
\end{table}

\subsection{Overall Performance}
To validate the effectiveness of our CSG framwork, we evaluate the generated cognitive structures in downstream task (i.e., KT and CD) to predict student performance. Tables \ref{tab:allKT} and \ref{tab:allCD} summarize the performance of our CSG-KT, CSG-CD, and all KT and CD baselines across four public datasets, where the average RMSE, ACC, and AUC values over 5-fold cross-validation, with the best results highlighted in bold. We observe and summarize that: \textbf{(\romannumeral 1)} As shown in Table \ref{tab:allKT}, CSG-KT not only significantly outperforms classical knowledge tracing models (e.g., DKT, SAKT) but also yields clear improvements over graph-based baselines that model only concept construction without capturing inter-concept relations (e.g., GKT, SKT, GRKT), and even surpasses recent SOTA methods (e.g., GRKT, MIKT, ENAS-KT). In Table \ref{tab:allCD}, CSG-CD significantly outperforms classical parameter-estimation based cognitive diagnosis models (e.g., IRT, MIRT), achieves substantial gains over neural cognitive diagnosis models that represent student ability solely at the concept dimension (e.g., NCD), and likewise exceeds the performance of heterogeneous student–concept–questions graph-based SOTA approaches. These results demonstrate that our generative cognitive structures not only provide a more comprehensive and accurate representation of student learning states but also reflect the dynamic evolution of cognitive structures over time. \textbf{(\romannumeral 2)} Despite substantial variations in scale and interaction density across the real-world datasets, CSG-KT and CSG-CD consistently exhibit robust and adaptable performance. More analysis of hyperparameters and inference time can be found in Appendix~\ref{app:hyperparameter} and Appendix~\ref{app:infer}.

\subsection{Ablation Study}

\begin{wraptable}{r}{0.33\textwidth}
  \centering
  \caption{Description of ablation variants of CSG.}
  \label{tab:des-variant}
  \resizebox{0.3\textwidth}{!}{
  \begin{tabular}{cccc}
    \toprule
    \multirow{2}{*}{\textbf{Variants}} & \multirow{2}{*}{\textbf{Pretraining}} & \multicolumn{2}{c}{\textbf{Optimization}} \\
    \cmidrule(lr){3-4}
     &  & \textbf{$r(\cdot)$} & \textbf{$r_{solo}(\cdot)$} \\
    \midrule
    Variant-1  & \ding{55} & \ding{55} & \ding{55} \\
    Variant-2  & \ding{51} & \ding{55} & \ding{55} \\
    Variant-3  & \ding{55} & \ding{51} & \ding{55} \\
    Variant-4  & \ding{55} & \ding{55} & \ding{51} \\
    Variant-5  & \ding{51} & \ding{51} & \ding{55} \\
    CSG      & \ding{51} & \ding{55} & \ding{51} \\
    \bottomrule
  \end{tabular}}
\vspace{-2ex}
\end{wraptable}
We evaluate several variants by comparing the prediction performance based on their sampled cognitive structures. We compare five variants (as shown in Table \ref{tab:des-variant}): \textbf{(\romannumeral 1)} Variant \(\boldsymbol{\mathrm{V}}_1\) directly uses rule-based simulated structures only; \textbf{(\romannumeral 2)} Variant \(\boldsymbol{\mathrm{V}}_2\) pretrains CSDPM on simulated cognitive structures only; \textbf{(\romannumeral 3)} Variant \(\boldsymbol{\mathrm{V}}_3\) skips pretraining and applies RL with generic reward $r(\cdot)$; \textbf{(\romannumeral 4)} Variant \(\boldsymbol{\mathrm{V}}_4\) skips pretraining and applies RL with SOLO-based reward $r_{solo}(\cdot)$; \textbf{(\romannumeral 5)} Variant \(\boldsymbol{\mathrm{V}}_5\) combines pretraining and RL with generic reward. The generic reward $r(\cdot)$ does not distinguish between SOLO levels while simply sums the matching functions $\mathcal M_{\mathcal V}$ and $\mathcal M_{\mathcal E}$ to produce a single scalar reward.

To ensure a fair comparison, we use the rule‐based simulated set $\tilde{\mathbb{G}}$ for variant \(\boldsymbol{\mathrm{V}}_1\), and sample the corresponding cognitive structures set $\mathbb{G}_0$ for each of variants \(\boldsymbol{\mathrm{V}}_2\)-\(\boldsymbol{\mathrm{V}}_5\). We then independently train and evaluate the downstream KT and CD models, denoted \(\boldsymbol{\mathrm{V}}_{i}\)\textbf{-KT} and \(\boldsymbol{\mathrm{V}}_{i}\)\textbf{-CD} for $i=1,\dots,5$.

Tables \ref{tab:ablKT} and \ref{tab:ablCD} present the ablation results for CSG-KT and CSG-CD, as well as their variant models, across four public datasets. We observe and summarize that: \textbf{(\romannumeral 1)} Overall, performance steadily improves from the simplest variant \(\boldsymbol{\mathrm{V}}_1\) through \(\boldsymbol{\mathrm{V}}_5\) to our CSG for both KT and CD tasks. \textbf{(\romannumeral 2)} For variant \(\boldsymbol{\mathrm{V}}_1\), performance nearly matches or even slightly surpasses classical baselines (e.g., DKT in KT and IRT, NCD in CD), demonstrating the effectiveness of our rule-based cognitive structure simulation. The embedded educational priors provide a reasonable approximation of student learning states. On Math1, Math2, and FrcSub where interaction sequences are short but item coverage is high, the simulation remains effective; on NIPS34, the longer interaction sequences compensate for lower coverage, yielding similarly strong results. \textbf{(\romannumeral 3)} Variant \(\boldsymbol{\mathrm{V}}_3\) generally outperforms \(\boldsymbol{\mathrm{V}}_2\) by a margin, indicating that task-driven optimization can uncover some patterns that similar to the educational priors in students’ learning processes and embed them into the sampled cognitive structures. \textbf{(\romannumeral 4)} The improvements of \(\boldsymbol{\mathrm{V}}_4\) over \(\boldsymbol{\mathrm{V}}_3\), and of the full CSG over \(\boldsymbol{\mathrm{V}}_5\), both demonstrate that modeling cognitive structure evolution via developmental levels is meaningful and validate the effectiveness of our fine-grained hierarchical solo-based reward signals.
\begin{table}[t]
\caption{Ablation study between CSG-KT and various variants for KT on different datasets.}
\label{tab:ablKT}
\centering
\resizebox{\linewidth}{!}{
\begin{tabular}{c|ccc|ccc|ccc|ccc}
\toprule
\textbf{Datasets}   & \multicolumn{3}{c|}{\textbf{Math1}} & \multicolumn{3}{c|}{\textbf{Math2}} & \multicolumn{3}{c|}{\textbf{FrcSub}}  & \multicolumn{3}{c}{\textbf{NIPS}} \\
\midrule
\textbf{Metrics}    & \textbf{AUC}\(\uparrow\) & \textbf{ACC}\(\uparrow\) & \textbf{RMSE}\(\downarrow\) 
 & \textbf{AUC}\(\uparrow\) & \textbf{ACC}\(\uparrow\) & \textbf{RMSE}\(\downarrow\)
 & \textbf{AUC}\(\uparrow\) & \textbf{ACC}\(\uparrow\) & \textbf{RMSE}\(\downarrow\)
 & \textbf{AUC}\(\uparrow\) & \textbf{ACC}\(\uparrow\) & \textbf{RMSE}\(\downarrow\)\\ 
\midrule
\textbf{\(\boldsymbol{\mathrm{V}}_1\)-KT} &  0.7742 &  0.6950 &  0.4596 &  0.7276 &  0.6745 &  0.4571 &  0.7944 &  0.7286 &  0.3955 &  0.6707 &  0.6204 &  0.4697 \\
\textbf{\(\boldsymbol{\mathrm{V}}_2\)-KT} &  0.7891 &  0.6996 &  0.4533 &  0.7421 &  0.6887 &  0.4543 &  0.8088 &  0.7430 &  0.3797 &  0.6751 &  0.6347 &  0.4674 \\
\textbf{\(\boldsymbol{\mathrm{V}}_3\)-KT} &  0.7942 &  0.7143 &  0.4472 &  0.7567 &  0.6930 &  0.4511 &  0.8233 &  0.7575 &  0.3641 &  0.6896 &  0.6491 &  0.4663 \\
\textbf{\(\boldsymbol{\mathrm{V}}_4\)-KT} &  0.7985 &  0.7191 &  0.4413 &  0.7614 &  0.6974 &  0.4491 &  0.8379 &  0.7621 &  0.3487 &  0.7042 &  0.6537 &  0.4604 \\
\textbf{\(\boldsymbol{\mathrm{V}}_5\)-KT} &  0.8047 &  0.7239 &  0.4356 &  0.7762 &  0.7018 &  0.4443 &  0.8426 &  0.7768 &  0.3335 &  0.7189 &  0.6583 &  0.4597 \\
\midrule
\textbf{CSG-KT}   & \textbf{0.8192}    & \textbf{0.7389}    & \textbf{0.4301}    & \textbf{0.7813}    & \textbf{0.7163}    & \textbf{0.4397 }   & \textbf{0.8575}    & \textbf{0.7917}    & \textbf{0.3186 }   & \textbf{0.7339}    & \textbf{0.6732}    & \textbf{0.4542} \\ 
\bottomrule 
\end{tabular}}
\end{table}

\begin{table}[t]
\caption{Ablation study between CSG-CD and various variants for CD on different datasets.}
\label{tab:ablCD}
\centering
\resizebox{\linewidth}{!}{
\begin{tabular}{c|ccc|ccc|ccc|ccc}
\toprule
\textbf{Datasets}   & \multicolumn{3}{c|}{\textbf{Math1}} & \multicolumn{3}{c|}{\textbf{Math2}} & \multicolumn{3}{c|}{\textbf{FrcSub}}  & \multicolumn{3}{c}{\textbf{NIPS}} \\
\midrule
\textbf{Metrics}    & \textbf{AUC}\(\uparrow\) & \textbf{ACC}\(\uparrow\) & \textbf{RMSE}\(\downarrow\) 
 & \textbf{AUC}\(\uparrow\) & \textbf{ACC}\(\uparrow\) & \textbf{RMSE}\(\downarrow\)
 & \textbf{AUC}\(\uparrow\) & \textbf{ACC}\(\uparrow\) & \textbf{RMSE}\(\downarrow\)
 & \textbf{AUC}\(\uparrow\) & \textbf{ACC}\(\uparrow\) & \textbf{RMSE}\(\downarrow\)\\ 
\midrule
\textbf{\(\boldsymbol{\mathrm{V}}_1\)-CD} &  0.7570 &  0.7377 &  0.4218 &  0.7667 &  0.7177 &  0.4508 &  0.8210 &  0.7963 &  0.3475 &  0.7471 &  0.6968 &  0.4511 \\
\textbf{\(\boldsymbol{\mathrm{V}}_2\)-CD} &  0.7613 &  0.7420 &  0.4157 &  0.7708 &  0.7219 &  0.4471 &  0.8354 &  0.8038 &  0.3309 &  0.7613 &  0.7010 &  0.4471 \\
\textbf{\(\boldsymbol{\mathrm{V}}_3\)-CD} &  0.7758 &  0.7465 &  0.4098 &  0.7851 &  0.7363 &  0.4406 &  0.8401 &  0.8085 &  0.3246 &  0.7757 &  0.7154 &  0.4413 \\
\textbf{\(\boldsymbol{\mathrm{V}}_4\)-CD} &  0.7805 &  0.7512 &  0.4041 &  0.7986 &  0.7409 &  0.4395 &  0.8550 &  0.8134 &  0.3235 &  0.7803 &  0.7200 &  0.4357 \\
\textbf{\(\boldsymbol{\mathrm{V}}_5\)-CD} &  0.7953 &  0.7661 &  0.4016 &  0.8043 &  0.7557 &  0.4352 &  0.8602 &  0.8285 &  0.3197 &  0.7951 &  0.7348 &  0.4303 \\
\midrule
\textbf{CSG-CD}   & \textbf{0.8104}    & \textbf{0.7812}    & \textbf{0.3933}    & \textbf{0.8193}    & \textbf{0.7608}    & \textbf{0.4301}    & \textbf{0.8757}    & \textbf{0.8439}    & \textbf{0.3172}    & \textbf{0.8002}    & \textbf{0.7499}    & \textbf{0.4251}   \\
\bottomrule
\end{tabular}}
\end{table}


\subsection{Visualization and Interpretability Analysis}
As shown in Fig.\ref{fig:caseall} of Appendix \ref{sec:casestudy}, we found that: \textbf{(\romannumeral 1)} Subfigure (a) shows the cognitive structure generated by CSG‐CD for student $s_5$ immediately before answering question $q_1$ (assessing concepts $k_{0,2,5,7,9}$); the student exhibits poor construction of both the individual concepts and their inter-concept relations, so CSG-CD predicts an incorrect response. Subfigure (b) shows the cognitive structure for student $s_{18}$ before the same question $q_1$; here, the student has good construction of all five tested concepts but still poor construction of their inter-concept relations, so CSG-CD again predicts an incorrect response. Subfigure (c) shows the cognitive structure for student $s_{37}$ before $q_1$; in this case, the student has good construction of both the tested concepts and their relations, so CSG-CD predicts a correct response. \textbf{(\romannumeral 2)} Subfigure (d) shows five representative cognitive structures generated by CSG-KT for student $s_{15}$ at different time points in their response sequence. Over the learning process, $s_{15}$’s cognitive structure evolves from almost no construction to a fully construction that successfully integrated entire knowledge system in $s_{15}$’s mind, broadly aligning with the cognitive development levels of SOLO taxonomy. The analysis demonstrates that the cognitive structures generated by CSG both reflect students’ subjective construction of the objective knowledge system and trace its evolution over learning. The results conform to established cognitive developments in educational psychology, thereby providing meaningful explanations for students’ response behavior.

\section{Limitations and Conclusion}\label{sec:conclusion}
CSG leverages diffusion models, which are generally more computationally intensive than classical architectures used in knowledge tracing and cognitive diagnosis, such as LSTMs and GNNs. However, recent advances in accelerating the denoising process of diffusion models~\cite{nichol2021improved,liu2022pseudo,song2023consistency,yin2024one} offer promising avenues to improve efficiency. Moreover, student cognitive structures typically do not require real-time updates, making the added computational cost acceptable in practical settings.

Besides, while we acknowledge that this rule-based simulation simplifies the inherently complex and nonlinear nature of cognitive development, it provides a crucial foundation for the first stage of our two-step training process. Inspired by LLM pretraining, we first learn initial representations using these rule-generated graphs, then refine them through reinforcement learning with a SOLO-taxonomy–based reward. This second stage allows the model to move beyond handcrafted assumptions and better capture authentic patterns of cognitive construction.

In summary, we introduced Cognitive Structure Generation (CSG), a novel framework for modeling students’ evolving cognitive structures through a diffusion model pretrained on educational priors and optimized with SOLO-based rewards. This two-stage approach enables the generation of interpretable, development-aligned cognitive structures. Empirical results across multiple benchmarks demonstrate that CSG substantially improves performance on knowledge tracing and cognitive diagnosis tasks. Our work underscores the importance of holistic cognitive structure modeling and offers a promising direction for interpretable and effective student modeling.



\bibliographystyle{unsrtnat}
\bibliography{mybibfile}

\input{appendix}




\end{document}

%% file: appendix.tex
\appendix

\section{The complete procedure of pretraining CSDPM}\label{sec:pretrain-csdpm}
\begin{algorithm}[H]
\caption{Pretraining CSDPM}
\label{alg:pretrain-csdpm}
\KwIn{Simulated dataset $\tilde{\mathbb{G}}$, diffusion steps $T$, loss weight $\lambda_{ve}$}
\While{not converged}{
  Sample $(\mathcal{G}_0, X^{T'}) \sim \tilde{\mathbb{G}}$\;
  \tcp{Sample a simulated cognitive structure and its interaction sequence}
  Sample $t \sim \mathcal{U}\llbracket 1, T\rrbracket$\;
  Sample $\mathcal{G}_t \sim q(\,\mathcal{G}_t |\mathcal{G}_0\,)$\;
  $z \gets f(\mathcal{G}_t, t)$ \tcp*{Graph-theoretic features}
  $h \gets \mathrm{emb}(X^{T'})$         \tcp*{Interaction-guidance features}
  $(\hat p^{\mathcal V},\,\hat p^{\mathcal E}) 
    \gets \phi_\theta(\mathcal{G}_t,\,z,\,h)$ \tcp*{Denoising pass}
  optimizer.step \((\mathcal L_{CE}({\hat p}^{\mathcal V},\mathcal{V}_0)+\lambda_{ve} \mathcal L_{CE}({\hat p}^{\mathcal E},\mathcal{E}_0))\) \tcp*{Cross-entropy loss}
}
\end{algorithm}

\section{The complete procedure of Policy Optimization}\label{sec:csdpm-policy-opt}
\begin{algorithm}[H]
\caption{Cognitive Structure Diffusion Policy Optimization}
\label{alg:csdpm-policy-opt}
\KwIn{Pretrained CSDPM $p_\theta$, diffusion steps $T$, reward function $r_{solo}(\cdot)$, learning rate $\eta$, number of trajectories $|\mathcal{D}|$, timestep samples $|\mathcal T|$, training steps $N$}
\KwOut{Optimized CSDPM $p_\theta$}

\For{$n=1,\dots,N$}{
  \For{$d=1,\dots,|\mathcal{D}|$}{
    Sample cognitive structure trajectory $\mathcal{G}_{0:T}^{(d)} \sim p_{\theta}(\mathcal{G}_{0:T})$\;
    Compute reward $r_{solo}(\mathcal{G}_0^{(d)})$ \;
    Sample random timesteps subset $\mathcal{T}_d \subseteq \llbracket 1,T\rrbracket$\;
  }
  \tcp{Estimate reward statistics}
  \quad$\bar{r}\leftarrow\frac{1}{|\mathcal D|}\sum_{d=1}^{|\mathcal D|}r_{solo}(\mathcal G_0^{(d)})$, \quad $\text{std}[r]\leftarrow\sqrt{\frac{1}{|\mathcal D|-1}\sum_{d=1}^{|\mathcal D|}(r_{solo}(\mathcal G_0^{(d)})-\bar{r})^2}$\;
  \tcp{Estimate eager policy gradient}
  \quad$\nabla_\theta J_{\mathrm{RL}}(\theta)\leftarrow\frac{1}{|\mathcal D|}\sum_{d=1}^{|\mathcal D|}\frac{T}{|\mathcal T_d|}\sum_{t\in\mathcal T_d}\frac{r_{solo}(\mathcal G_0^{(d)})-\bar{r}}{\text{std}[r]}\nabla_{\theta}\log p_{\theta}(\mathcal G_0^{(d)}|\mathcal G_t^{(d)})$\;
  \tcp{Update parameters}
  \quad$\theta\leftarrow\theta+\eta\cdot \nabla_\theta J_{\mathrm{RL}}(\theta)$\;
}
\end{algorithm}

\section{Statistics of all four datasets.}\label{sec:statisdata}
\begin{table}[H]
\caption{Statistics of all four datasets.}
\label{tab:datasets}
\centering
\resizebox{0.7\linewidth}{!}{
\begin{tabular}{lllll}
\toprule
\textbf{Datasets} & \textbf{Math1} & \textbf{Math2}   & \textbf{FrcSub}    & \textbf{NIPS34} \\ \midrule
\# of students     & 4,209           & 3,911  & 536      & 4918    \\
\# of exercises     & 20           & 20     & 20       & 948   \\
\# of knowledge concepts    & 11             & 16     & 8        & 57     \\
\# of interactions   & 72,359         & 78,221 & 10,720 & 1,399,470 \\
\# of interactions per student   & 17.19         & 20.00 & 20.00 & 284.56 \\
\bottomrule
\end{tabular}
}
\end{table}

\section{Implementation Details}\label{sec:imp}
In our experiments, all interaction data of students are randomly split with a ratio of 8:1:1 to create the training and test sets at the level of interaction records. No test interaction appears in the training set, and no model is ever trained on test data. For the test set, CSG generates cognitive structures from a student's interaction history up to time \(T\). For KT, this structure is used to predict performance at \(T+1\). For CD, the model never observes the response \(r_{ij}\) for the item it is asked to predict.

For the parameterization of the CSDPM, we employ the extended Graph Transformer architecture from \cite{DBLP:journals/corr/abs-2012-09699,DBLP:conf/iclr/VignacKSWCF23}, configuring it with $8$ transformer layers, whose hidden dimensions (e.g., MLP, attention heads, and feed-forward layers) are set identically to those in \cite{DBLP:conf/iclr/VignacKSWCF23}. For pretraining the CSDPM, the CSDPM is trained using a uniform transition kernel for diffusion and the AdamW optimizer, with the number of diffusion steps $T$ set as $500$, node–edge loss balancing coefficient $\lambda_{ve}$ $(0,1)$, the batch size $(64,512)$, dropout rate \((0,0.5)\), and initial learning rate [1e-5,1e-2] with weight decay tuned via random or grid search strategy.
The number of sampled trajectories $\mathcal{D}$ is searched in $\{128,256,512\}$. For CSG-KT and CSG-CD, the dimension of the graph pooling for cognitive state representation is searched in \(\{8,16,32,64\}\). To configure the training process, we initialize the parameters using Xavier initialization \citep{glorot2010understanding} and employ flexible methods such as random, grid, and bayes search{\normalfont\&}select strategies. For fairness, The hyper-parameter settings of the baseline models are have been further tuned using the same tuning strategies to achieve optimal results. All experiments were run on Linux servers equipped with an Intel Xeon Platinum 8352V CPU and NVIDIA RTX 4090 GPUs. 


\section{Additional Analysis on Hyperparameters}\label{app:hyperparameter}
We conducted a sensitivity analysis of some key parameters. We summarize the following observations and conclusions: The optimal node–edge loss balancing coefficient $\lambda_{ve}\in(0,1)$ was 0.5 for Math1, Math2, and FrcSub, and 0.6 for NIPS34, that has larger number of nodes yields a correspondingly greater number of edges. For both CSG-KT and CSG-CD, the optimal graph pooling dimension was 16 for Math1 and Math2, 8 for FrcSub, and 32 for NIPS34. 

We also explored different scaling schemes—including both linear and exponential reward progressions early on and found them to be less optimal in downstream performance and training stability. The optimal SOLO reward tuple $(r_1,r_2,r_3,r_4,r_5)$ is found to be $(0,2,12,32,36)$ for both CSG-KT and CSG-CD across all datasets. we designed them based on two considerations. First, our preliminary experiments showed that the distribution of levels among generated structures tends to be approximately bell-shaped—most samples fall into the middle levels (uni-structural, multi-structural, and relational), while extreme levels (pre-structural and extended abstract) are relatively rare. Second, we aimed to encourage the model to reach higher alignment levels without introducing instability into the training process. To this end, we adopted a progressively increasing reward scale that assigns smaller gains between lower levels and larger gains toward the upper levels. This asymmetry creates a learning signal that meaningfully distinguishes early-stage from advanced cognitive alignment. At the same time, we avoided overly steep jumps between adjacent levels to prevent reward explosion and reduce the risk of high-variance gradients during reinforcement learning. In practice, the selected reward tuple yields stable training behavior and consistent performance across datasets, and we apply the same values throughout all experiments without dataset-specific tuning.

\section{Inference Time Analysis}\label{app:infer}

We further report the inference time of CSG for generating a single cognitive structure graph. As shown in Table~\ref{tab:inference_time}, the inference time remains low across datasets of different sizes, 
demonstrating the practical feasibility and efficiency of our CSG.
\begin{table}[H]
\caption{Inference time for generating a single cognitive structure graph.}
\centering
\begin{tabular}{ccc}
\hline
\textbf{Dataset} & \textbf{Nodes} & \textbf{Inference Time (ms)} \\
\hline
Math1   & 11 & 2.61  \\
Math2   & 16 & 4.24  \\
FrcSub  & 8  & 0.74  \\
NIPS34    & 57 & 25.65 \\
\hline
\end{tabular}
\label{tab:inference_time}
\end{table}

\section{Example of the calculation process of Cognitive Structure Simulation}\label{app:cal}

Given five questions \(q_1\)–\(q_5\) that assess the concepts \textit{Sine Theorem} and \textit{Cosine Theorem}, we make an idealized assumption: if a question involves only one concept, its weight for that concept is set to 1; if it involves both concepts, the weights for each concept are set to 0.5. Suppose a student’s responses to these five questions at timestamp \(T\) are recorded, as shown in the Table \ref{tab:question_weights} below.
\begin{table}[H]
\caption{Example of question weights and student responses.}
\centering
\begin{tabular}{cccc}
\hline
\textbf{Question} & \textbf{Sine Weight} & \textbf{Cosine Weight} & \textbf{Response} \\
\hline
$q_1$ & 1.0 & 0.0 & Correct \\
$q_2$ & 1.0 & 0.0 & Correct \\
$q_3$ & 0.5 & 0.5 & Correct \\
$q_4$ & 0.5 & 0.5 & Incorrect \\
$q_5$ & 1.0 & 0.0 & Incorrect \\
\hline
\end{tabular}
\label{tab:question_weights}
\end{table}

Accordingly, using Eqs.\ref{eq:uoc} and \ref{eq:uor}, we can calculate the student’s construction for the concepts \textit{Sine Theorem} and \textit{Cosine Theorem}, the node-level term \(v^T(\text{Sine Theorem})\) and the edge-level term \(e^T(\text{Sine–Cosine Theorem})\) in the simulated cognitive structure, as follows:
\[v^T(\text{Sine Theorem}) = \frac{1 + 1 + 0.5 + 0 + 0}{1 + 1 + 0.5 + 0.5 + 1} = \frac{2.5}{4.0} = 0.625,\]
\[
e^T(\text{Sine–Cosine Theorem}) = \frac{0 + 0 + 0.5 + 0 + 0}{0 + 0 + 0.5 + 0.5 + 0} = \frac{0.5}{1.0} = 0.5.
\]

\section{Case Study of CSG} \label{sec:casestudy}
\begin{figure}[H]
        \centering
        \includegraphics[width=\linewidth]{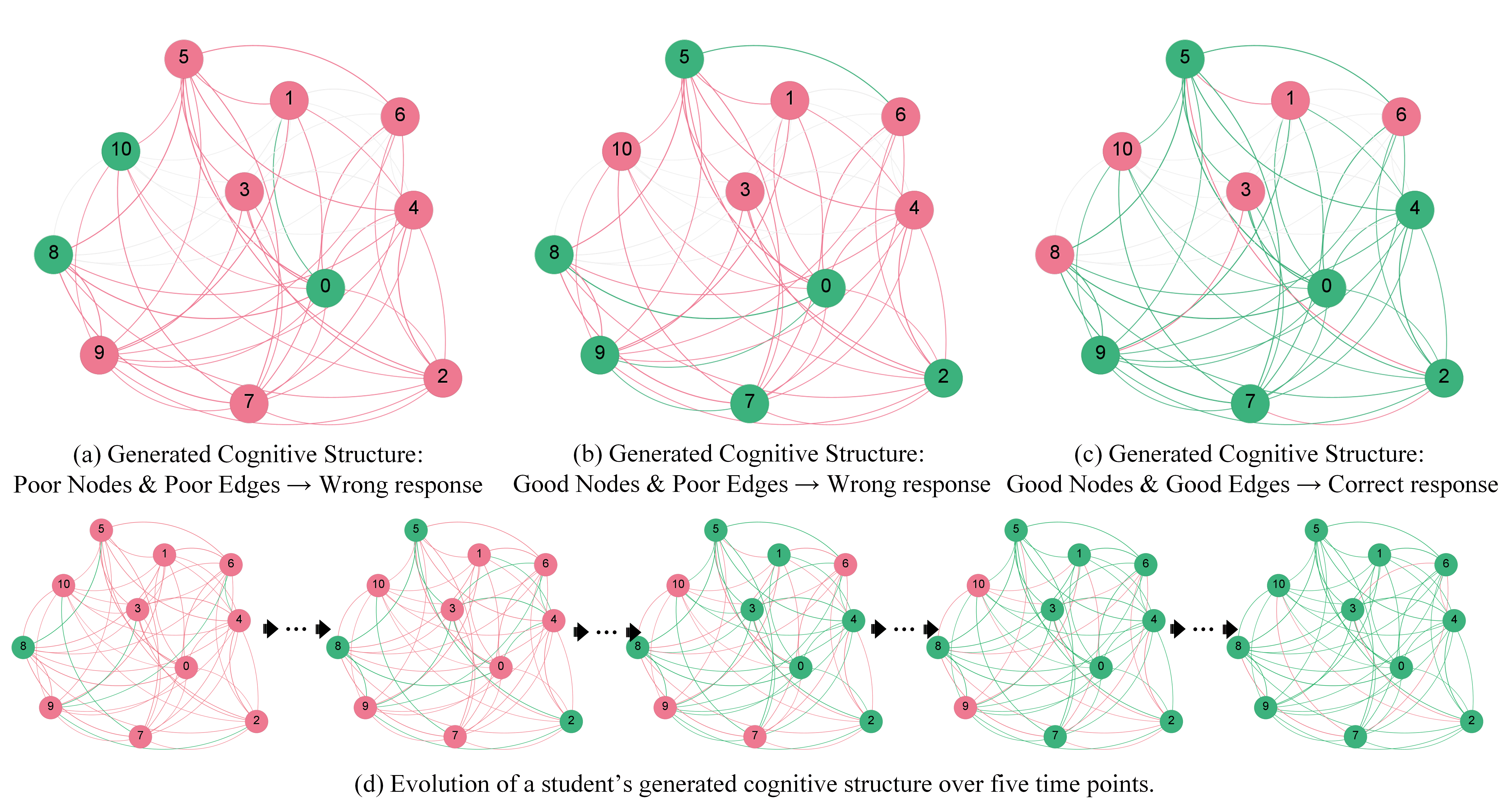}
        \caption{\textbf{(Case).} Examples of generated cognitive structures and the evolution process. Each graph depicts a student’s generated cognitive structure at a given timestamp. Nodes represent the student’s construction of concepts, while edges represent their construction of inter-concept relations. Green indicates fully constructed elements, red indicates elements not yet constructed, and gray denotes low-frequency or irrelevant edges shown for clarity.}
        \label{fig:caseall}
\end{figure}

\begin{table}[H]
\caption{List of knowledge concepts in Math1.}
\centering
\begin{tabular}{cl}
\hline
\textbf{No.} & \textbf{Concept Name} \\
\hline
0  & Set \\
1  & Inequality \\
2  & Trigonometric function \\
3  & Logarithm versus exponential \\
4  & Plane vector \\
5  & Property of function \\
6  & Image of function \\
7  & Spatial imagination \\
8  & Abstract summarization \\
9 & Reasoning and demonstration \\
10 & Calculation \\
\hline
\end{tabular}
\label{tab:skill_list}
\end{table}